\documentclass{article}
\usepackage{ijcai23}

\usepackage{times}
\usepackage{soul}
\usepackage{url}
\usepackage[hidelinks]{hyperref}
\usepackage[utf8]{inputenc}
\usepackage[small]{caption}
\usepackage{graphicx}
\usepackage{amsmath}
\usepackage{amsthm}
\usepackage{amsmath}
\usepackage{amssymb}
\usepackage{booktabs}
\usepackage{algorithm}
\usepackage[switch]{lineno}
\usepackage{color}
\usepackage{multirow}
\usepackage{array}
\usepackage{subcaption}

\newcommand{\bigname}[0]{\textsc{PowerBEV}}
\newcommand{\name}[0]{PowerBEV}
\newcolumntype{L}[1]{>{\raggedright\let\newline\\\arraybackslash\hspace{0pt}}m{#1}}
\newcolumntype{C}[1]{>{\centering\let\newline\\\arraybackslash\hspace{0pt}}m{#1}}
\newcolumntype{R}[1]{>{\raggedleft\let\newline\\\arraybackslash\hspace{0pt}}m{#1}}

\def\secref#1{Section~\ref{#1}}
\def\figref#1{Figure~\ref{#1}}
\def\tabref#1{Table~\ref{#1}}
\def\eqref#1{Equation~(\ref{#1})}

\title{PowerBEV: A Powerful Yet Lightweight Framework for \\
 Instance Prediction in Bird’s-Eye View}

\author{
Peizheng Li$^{*,1,3}\!$
\and
Shuxiao Ding$^{*,1,2}$\!\and
Xieyuanli Chen$^{2}$\!\and
Niklas Hanselmann$^{1,3}$\!\and
\\
Marius Cordts$^1$\!\And
Juergen Gall$^{2,4}$
\affiliations
$^1$Mercedes-Benz AG, Stuttgart, Germany \\
$^2$University of Bonn, Bonn, Germany \\
$^3$University of Tübingen, Tübingen, Germany \\
$^4$Lamarr Institute for Machine Learning and Artificial Intelligence, Germany
\emails
\{peizheng.li, shuxiao.ding, niklas.hanselmann, marius.cordts\}@mercedes-benz.com, \\
xieyuanli.chen@nudt.edu.cn, gall@iai.uni-bonn.de
}

\begin{document}

\maketitle

\begin{abstract}
	Accurately perceiving instances and predicting their future motion are key tasks for autonomous vehicles, enabling them to navigate safely in complex urban traffic.
	While bird's-eye view (BEV) representations are commonplace in perception for autonomous driving, their potential in a motion prediction setting is less explored.
	Existing approaches for BEV instance prediction from surround cameras rely on a multi-task auto-regressive setup coupled with complex post-processing to predict future instances in a spatio-temporally consistent manner.
	In this paper, we depart from this paradigm and propose an efficient novel end-to-end framework named \bigname{}, which differs in several design choices aimed at reducing the inherent redundancy in previous methods.
	First, rather than predicting the future in an auto-regressive fashion, \bigname{} uses a parallel, multi-scale module built from lightweight 2D convolutional networks.
	Second, we show that segmentation and centripetal backward flow are sufficient for prediction, simplifying previous multi-task objectives by eliminating redundant output modalities.
	Building on this output representation, we propose a simple, flow warping-based post-processing approach which produces more stable instance associations across time.
	Through this lightweight yet powerful design, \bigname{} outperforms state-of-the-art baselines on the NuScenes Dataset and poses an alternative paradigm for BEV instance prediction.
	We made our code publicly available at: \url{https://github.com/EdwardLeeLPZ/PowerBEV}.
\end{abstract}

\def\thefootnote{*}\footnotetext{Equal contribution.}

%%%%%%%%%%%%%%%%%%%%%%%%%%%%%%%%%%%%%%%%%%%%%%%%%%%%%%%%%%%%%%%%%%%%%%%%%%%%%%%%
\section{Introduction}
\label{sec:intro}

\begin{figure}[t]
	\centering
	\includegraphics[width=1.0\linewidth]{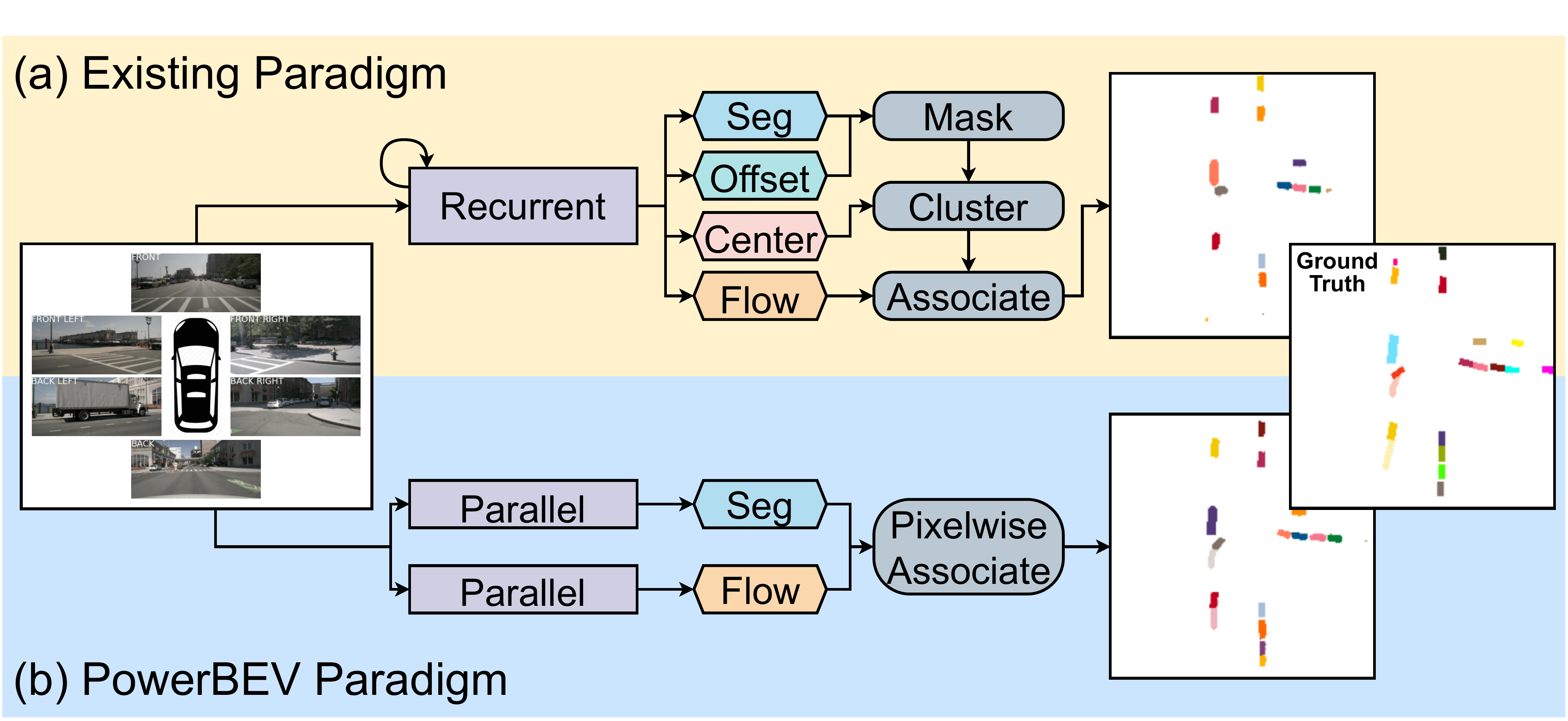}
	\caption{\textbf{\name{} vs.\ Existing Paradigm:} The existing prediction paradigm (a) outputs 4 predictions per frame using spatial RNNs. After masking out background grid cells, instance ID assignment is performed by grid clustering, followed by instance-level association. To eliminate the framework redundancy, we propose a more lightweight yet powerful parallel prediction paradigm, namely PowerBEV (b). It consists entirely of 2D CNNs supplemented by flow warping post-processing based on only 2 outputs.}
	\vspace{-0.2cm}
	\label{fig:motivation}
\end{figure}

Accurately acquiring surrounding vehicle information is a key challenge for autonomous driving systems.
In addition to the precise detection and localization of road users at present, predicting their future motion is also of great importance, considering the high complexity and dynamics of the driving environment.
A widely accepted paradigm is to decouple these tasks into separate modules.
Under this paradigm, objects of interest are first detected and localized through sophisticated perception models and associated across multiple frames.
Then, the past motion of these detected objects is used to forecast their potential future movement via parametric trajectory models~\cite{Luo2018FastAF,Liang2020PnPNetEP}.
However, by separating the perception and the motion model for forecasting, the whole system is prone to errors in the first stage.

In recent years, many works have demonstrated the potential of the bird's-eye view (BEV) representation for accurate vision-centric driving environment perception~\cite{philion2020lift,huang2021bevdet,li2022bevformer}.
To solve the error accumulation problem, researchers seek to exploit end-to-end frameworks to determine object locations directly in the BEV and forecast global scene changes in the form of an occupancy grid map~\cite{hu2021fiery,akan2022stretchbev,zhang2022beverse}.
Although employing the end-to-end paradigm, existing approaches forecast multiple, partially redundant representations like segmentation map, instance centers, forward flow, and offsets pointing to instance centers as shown in Figs.~\ref{fig:motivation} and \ref{fig:similarities_in_multi-tasks}.
These redundant representations require not only various loss terms, but also complex post-processing to obtain instance predictions.

In this work, we simplify the multi-task setting used in previous works and propose an approach that requires only two output modalities: segmentation maps and flow.
Specifically, we compute instance centers directly from the segmentation, allowing the omission of a redundant separate center map.
This additionally eliminates the potential for inconsistencies between the estimated centers and predicted segmentation.
Furthermore, contrary to the forward flow used in previous works, we compute a centripetal backward flow.
This is a vector field pointing from each occupied pixel at present to its corresponding instance center in the previous frame.
It combines the pixel- and instance-level association into a single pixel-wise instance assignment task.
Thus, the offset head is no longer required.
Morever, this design choice simplifies the association since it no longer needs multiple steps.
Compared to auto-regressive models, we also find that 2D convolutional networks are sufficient for the proposed \bigname{} framework to obtain satisfactory instance predictions, which results in a lightweight yet powerful framework.

We evaluate our approach on the NuScenes dataset~\cite{Caesar2019nuScenesAM}, where our method outperforms existing frameworks and achieves state-of-the-art instance prediction performance.
We further perform ablation studies to validate the design of our powerful but lightweight framework.

Our main contributions can be summarized as follows:
\begin{itemize}
	\item We propose \bigname{}, a novel and elegant vision-based end-to-end framework that only consists of 2D-convolutional layers to perform perception and forecasting of multiple objects in BEVs. 
	\item We demonstrate that over-supervision caused by redundant representations impairs the forecasting capability. In contrast, our method accomplishes both semantic and instance-level agent prediction by simply forecasting segmentation and centripetal backward flow.
	\item The proposed assignment based on centripetal backward flow is superior to the previous forward flow in combination with the traditional Hungarian Matching algorithm.
\end{itemize}

%%%%%%%%%%%%%%%%%%%%%%%%%%%%%%%%%%%%%%%%%%%%%%%%%%%%%%%%%%%%%%%%%%%%%%%%%%%%%%%%
\section{Related Work}
\label{sec:related}

\subsection{BEV for Camera-based 3D Perception}
While LiDAR-based perception approaches often map a 3D point cloud onto the BEV plane and perform BEV segmentation~\cite{Fei2021PillarSegNetPS,Peng2021MASSMS} or 3D bounding box regression~\cite{Yang2018PIXORR3,Lang2018PointPillarsFE,Yin2020Centerbased3O}, the transformation of a monocular camera image into a BEV representation remains an ill-posed problem.
Although there are methods~\cite{Fei2020SemanticVoxelsSF,liu2022bevfusion,dong2022superfusion,liang2022bevfusion} that combine LiDAR and camera data to generate BEVs, they rely on accurate multi-sensor calibration and synchronization.

LSS (Lift Splat Shoot)~\cite{philion2020lift} can be regarded as the first work that lifts 2D features to 3D and projects the lifted features onto the BEV plane. It discretizes the depth and predicts a distribution over depth. The image features are then scaled and  distributed across the depth dimension according to the distribution.
BEVDet~\cite{huang2021bevdet} adapts LSS to 3D object detection from BEV feature maps.
Tesla AI Day 2021~\cite{TesalAIDay2021} first proposes to use a Transformer architecture~\cite{Vaswani2017AttentionIA} to fuse multi-view camera features into BEV feature maps, where the cross-attention between dense BEV queries and perspective image features acts as the view transformation.
This approach is further improved by leveraging camera calibration and deformable attention~\cite{Zhu2020DeformableDD} in BEVFormer~\cite{li2022bevformer} and BEVSegFormer~\cite{Peng2022BEVSegFormerBE} to reduce the quadratic complexity of Transformers.
Furthermore, it has been shown that temporal modeling of BEV features achieves a significant performance improvement for 3D detection~\cite{li2022bevformer,Huang2022BEVDet4DET,jiang2022polarformer} at the cost of a high computation and memory consumption.
Unlike detection or segmentation, the forecasting task naturally needs temporal modeling of the historical information.
To tackle this, our approach extracts spatio-temporal information using a lightweight fully-convolutional network on top of LSS, which is both effective and efficient.

\subsection{BEV-based Future Prediction}
Early BEV-based prediction methods render the past trajectories into a BEV image and use CNNs to encode the rasterized input~\cite{bansal2018chauffeurnet,hong2019rules,chai2019multipath}, which assumes perfect detection and tracking of the objects.
Another line of works conducts end-to-end trajectory forecasting directly from LiDAR point clouds~\cite{casas2018intentnet,Luo2018FastAF,casas2020spagnn,Liang2020PnPNetEP}.
Unlike instance-level trajectory prediction, MotionNet~\cite{wu2020motionnet} and MP3~\cite{Casas2021MP3AU} tackle the forecasting task by a motion (flow) field for each occupancy grid.
In contrast to the above mentioned approaches that rely on LiDAR data, FIERY~\cite{hu2021fiery} first predicts a BEV instance segmentation solely from multi-view camera data.
FIERY extracts multi-frame BEV features following LSS~\cite{philion2020lift}, fuses them into a spatio-temporal state using a recurrent network, and then conducts a probabilistic instance prediction.
StretchBEV~\cite{akan2022stretchbev} improves FIERY using a stochastic temporal model with stochastic residual updates.
BEVerse~\cite{zhang2022beverse} proposes an iterative flow warping in latent space for prediction in a multi-task BEV perception framework.
These approaches follow Panoptic-DeepLab~\cite{Cheng2019PanopticDeepLabAS} that utilizes four different heads to compute a semantic segmentation map, instance centers, per-pixel centripetal offsets, and future flow.
They rely on complex post-processing to generate the final instance prediction from these four representations.
In this paper, we show that only two heads, namely semantic segmentation and centripetal backward flow, together with a simplified post-processing are sufficient for future instance prediction.

%%%%%%%%%%%%%%%%%%%%%%%%%%%%%%%%%%%%%%%%%%%%%%%%%%%%%%%%%%%%%%%%%%%%%%%%%%%%%%%%
\section{Approach}
\label{sec:main}

\begin{figure*}[t]
	\centering
	\includegraphics[width=0.90\linewidth]{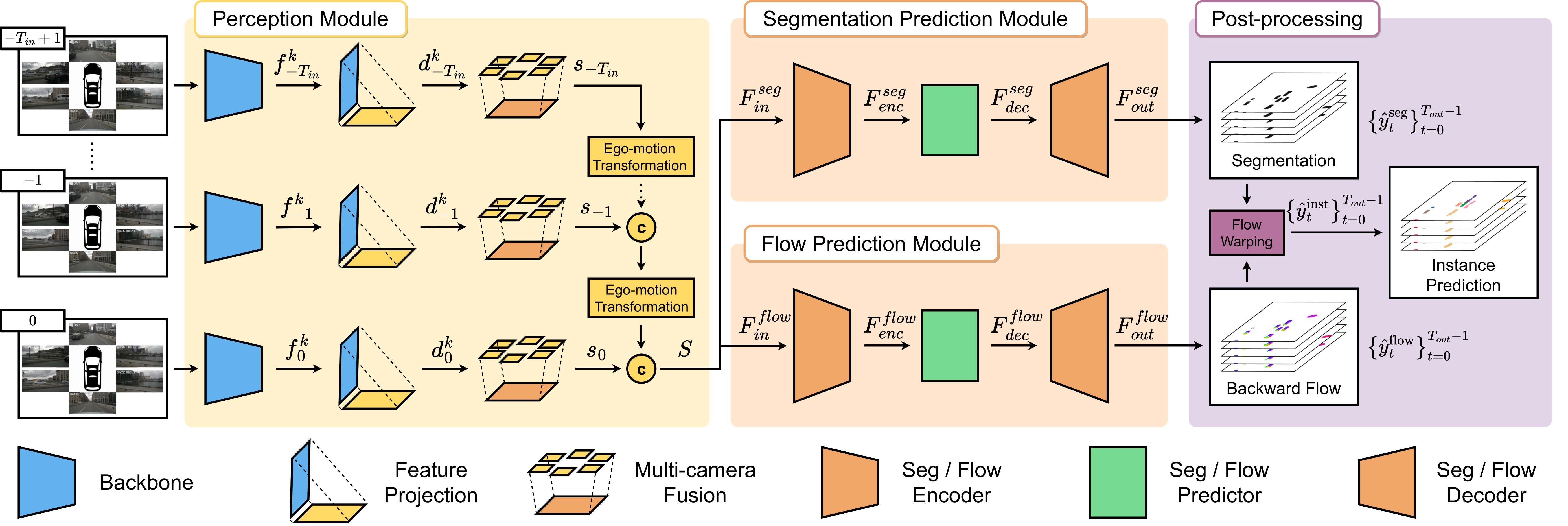}
	\caption{\textbf{Architecture of our Proposed End-to-End Framework:} In \name, the perspective features extracted by the perception module (yellow area) from surrounding camera images of each frame are projected into the BEV plane and then fused and stacked into the current global dynamic state. Subsequently, two independent prediction modules with the same structure (orange area) take the current state as input and predict the segmentation maps and centripetal backward flow for the future frames. Finally, future multi-frame instance predictions are generated by the flow warping post-processing (purple area).}
	\label{fig:framework_architecture}
	\vspace{-0.2cm}
\end{figure*}

In this section, we outline our proposed end-to-end framework. An overview of the approach is illustrated in~\figref{fig:framework_architecture}.
It consists of three main parts: a perception module, a prediction module and a post-processing stage.
The perception module follows LSS \cite{philion2020lift} and takes $M$ multi-view camera images for $T_\text{in}$ timestamps as input and lifts them into $T_\text{in}$ BEV feature maps (see~\secref{sec:perception}).
The prediction module then fuses the spatio-temporal information contained in the extracted BEV features (see~\secref{sec:msstc}) and predicts a sequence of segmentation maps and centripetal backward flows for $T_\text{out}$ future frames in parallel (see~\secref{sec:multitask}).
Finally, future instance predictions are recovered from the predicted segmentation and flow through a warping-based post-processing (see~\secref{sec:instance}).
In the following we describe each of the involved components in detail.

\subsection{LSS-based Perception Module}
\label{sec:perception}
To obtain visual features for prediction, we follow previous works~\cite{hu2021fiery,akan2022stretchbev} and build on LSS ~\cite{philion2020lift} to extract BEV feature grids from surround camera images.
More specifically, for each camera image $k \in \{1, \dots, 6\}$ at time $t$, we apply a shared EfficientNet~\cite{tan2019efficientnet} backbone to extract perspective features $f^{k}_{t}\in\mathbb{R}^{(C_{p} + D_{p}) \times H_{p} \times W_{p}}$,
where we designate the first $C_{p}$ channels of $f^{k}_{t}$ to represent a context feature $f^{k}_{t,C} \in \mathbb{R}^{C_{p} \times H_{p} \times W_{p}}$ and the following $D_p$ channels to represent a categorical depth distribution $f^{k}_{t,D} \in \mathbb{R}^{C_{p} \times H_{p} \times W_{p}}$.
A 3D feature tensor $d^{k}_{t}\in\mathbb{R}^{C_{p} \times D_{p} \times H_{p} \times W_{p}}$ is constructed by means of the outer product
\begin{equation}
	d^{k}_{t} = f^{k}_{t,C} \otimes  f^{k}_{t,D},
\end{equation}%\par
which represents a lifting of the context feature $f^{k}_{t,C}$ into different depths $D_p$ according to the estimated depth distribution confidence $f^{k}_{t,D}$.
Afterwards, the per-camera feature distribution maps $d^{k}_{t}$ at each timestamp are transformed to the ego-vehicle-centered coordinate system, leveraging known intrinsics and extrinsics of the corresponding cameras.
They are then weighted along the height dimension to obtain the global BEV state $s_{t}\in\mathbb{R}^{C_\text{in} \times H \times W}$ at timestamp $t$, where $C_\text{in}$ is the number of state channels and $(H, W)$ the grid size of the BEV state maps.
Finally, all BEV states $\left\{ s_{t} \right\}_{t=-T_\text{in}+1}^{0}$ are unified to the current frame and stacked as in FIERY~\cite{hu2021fiery}, thus representing current global dynamics $S\in\mathbb{R}^{C_\text{in} \times T_\text{in} \times H \times W}$ independent of ego-vehicle positions.

\subsection{Multi-scale Prediction Module}
\label{sec:msstc}
\begin{figure}[t]
	\centering
	\includegraphics[width=1.0\linewidth]{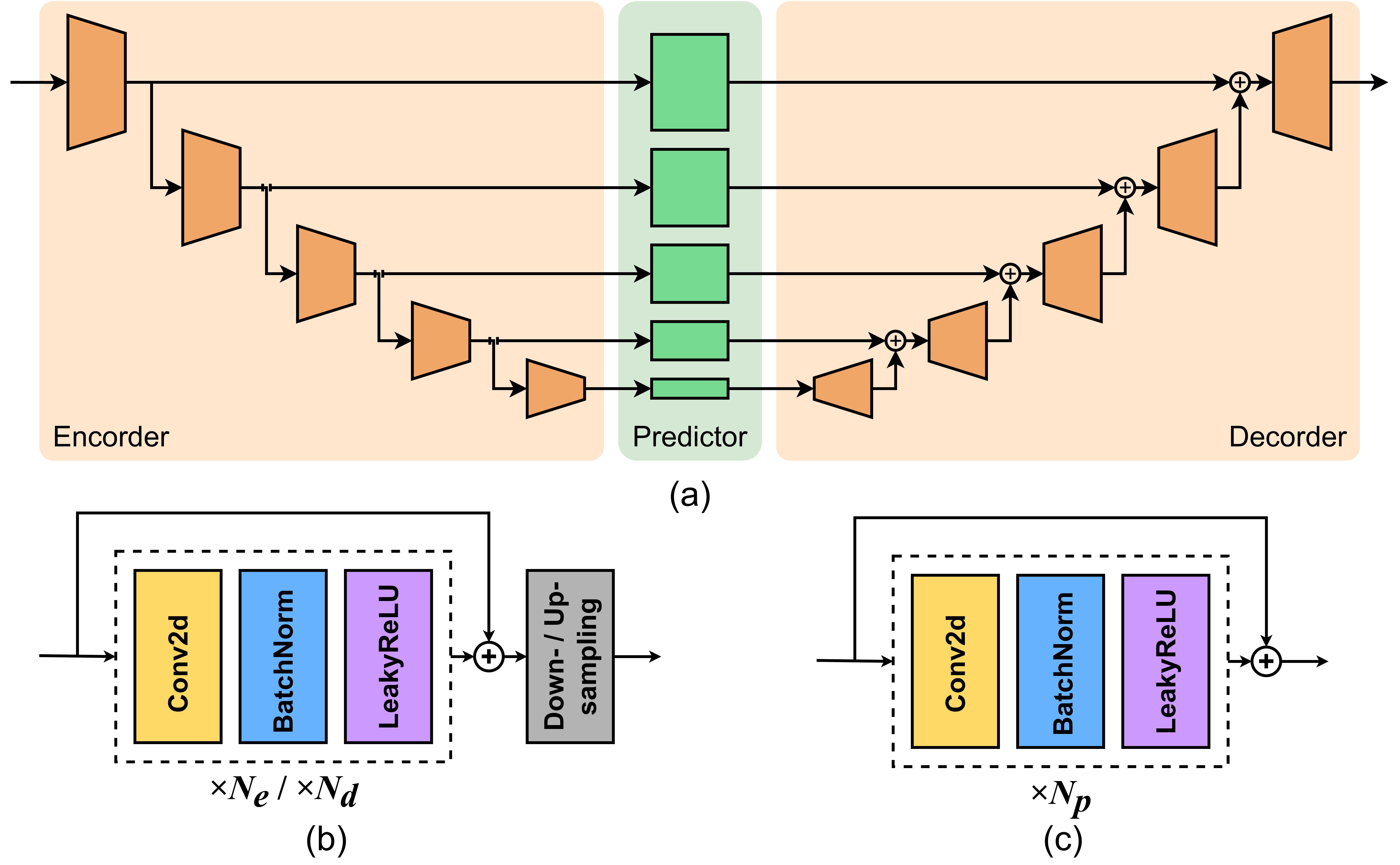}
	\caption{\textbf{Architecture of the Multi-scale Prediction Model:} (a) Overview; (b) Encoder/Decoder block with the down-/up-sampling layer; (c) Predictor block. $N_e$, $N_d$, $N_p$ denotes the number of encoder, decoder and predictor blocks respectively.}
	\label{fig:prediction_module_architecture}
	\vspace{-0.2cm}
\end{figure}

Having obtained a compact representation $S$ of the past context, we use a multi-scale U-Net-like encoder-decoder architecture that takes the observed BEV feature maps as input and predicts future segmentation maps and centripetal backward flow fields, as shown in \figref{fig:prediction_module_architecture}.
To achieve the spatio-temporal feature processing using only 2D convolutions, we collapse the time and feature dimensions into one single dimension, resulting in an input tensor $F_\text{in}\in\mathbb{R}^{(C_\text{in} \times T_\text{in}) \times H \times W}$.
The encoder first downsamples $F_\text{in}$ spatially step by step, producing multi-scale BEV features $F_\text{enc} \in\mathbb{R}^{(C_{i} \times T_\text{in}) \times \frac{H}{2^i} \times \frac{W}{2^i}}$, where $i \in \{1,\dots,5\}$.
In an intermediate predictor stage, the features are mapped from $C_{i} \times T_\text{in}$ to $C_{i} \times T_\text{out}$ to get $F_\text{dec} \in\mathbb{R}^{(C_{i} \times T_\text{out}) \times \frac{H}{2^i} \times \frac{W}{2^i}}$.
Finally, the decoder, which mirrors the encoder, reconstructs the future BEV features $F_\text{out}\in\mathbb{R}^{(C_\text{out} \times T_\text{out}) \times H \times W}$ at the original scale.

Each branch is supervised to predict future segmentation maps or centripetal backward flow fields, respectively.
Considering the differences in tasks and supervision, we use the same architecture for each branch but without weight-sharing.
Compared to previous work building on spatial LSTMs or spatial GRUs, our architecture leverages only 2D convolutions and largely alleviates the limitations of spatial RNNs in solving long-range temporal dependencies.

\subsection{Multi-task Settings}
\label{sec:multitask}
\begin{figure}[t]
	\centering
	\includegraphics[width=1.0\linewidth]{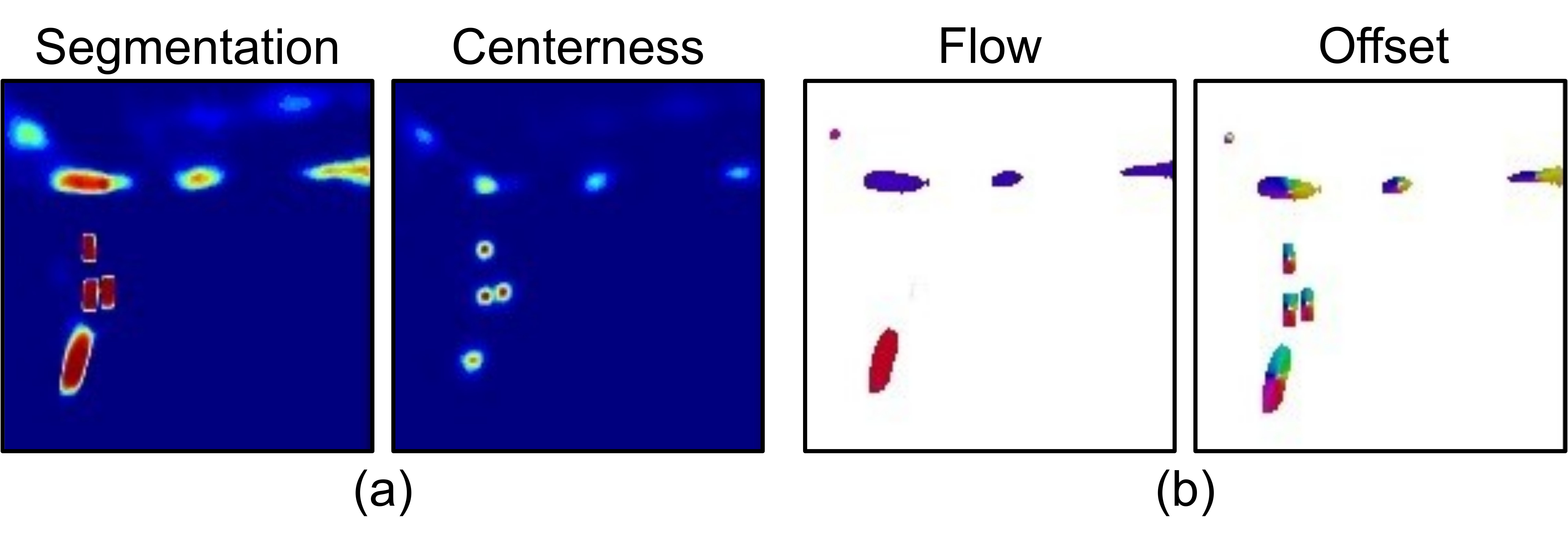}
	\vspace{-0.6cm}
	\caption{\textbf{Task Similarities:} (a) segmentation probability and centerness are both Gaussian distributions; (b) flow and offset are both regression tasks within occupied regions.}
	\vspace{-0.2cm}
	\label{fig:similarities_in_multi-tasks}
\end{figure}

Existing approaches~\cite{hu2021fiery,akan2022stretchbev,zhang2022beverse} follow a bottom-up pipeline~\cite{Cheng2019PanopticDeepLabAS} that generates instance segmentation for each frame and then associates instances across frames using Hungarian Matching (HM)~\cite{Kuhn1955TheHM} based on the forward flow.
Consequently, four different heads are required: semantic segmentation, centerness, future forward flow and per-pixel centripetal offsets in BEV (\figref{fig:motivation}.a).
This leads to model redundancy and instability due to multi-task training.

By comparison, we first find that both flow and centripetal offsets are regression tasks within the instance mask (\figref{fig:similarities_in_multi-tasks}.b) and the flow can be understood as the motion offset.
In addition, both quantities are combined with the centerness in two stages: (1) centripetal offset groups pixels to the predicted instance center in each frame to assign pixels to instance IDs; (2) flow is used to match the centers in two consecutive frames for instance ID association.
Based on the above analysis, it is intuitive to solve both tasks using a unified representation.

To this end, we propose the \textit{backward centripetal flow field}, which is the displacement vector from each foreground pixel at time $t$ to the object center of the associated instance identity at time $t-1$.
This unifies the pixel-to-pixel backward flow vector and the centripetal offset vector into a single representation.
Using our proposed flow, each occupied pixel can be directly associated to an instance ID in the previous frame.
This eliminates the need for an additional clustering step that assigns pixels to instances, simplifying the two-stage post-processing used in previous works into a single-stage association task (\figref{fig:motivation}.b).
This instance association mechanism is further discussed in detail in~\secref{sec:instance}.

Furthermore, we find that the predictions of the semantic segmentation map and the centerness reveals a very high similarity since the centerness essentially corresponds to the center positions of the semantic instances (\figref{fig:similarities_in_multi-tasks}.a).
Thus, we propose to directly infer the object centers by extracting the local maxima in the predicted segmentation map using the method by \cite{Zhou2019ObjectsAP}. This eliminates the need to seperately predict centerness.

To summarize, our network produces only two outputs: the semantic segmentation $\left\{ \hat{y}_{t}^\text{seg} \right\}_{t=0}^{T_\text{out}-1}$ and the backward centripetal flow $\left\{ \hat{y}_{t}^\text{flow} \right\}_{t=0}^{T_\text{out}-1}$.
We use top-$k$ cross-entropy \cite{Berrada2018SmoothLF} with $k=25\%$ as segmentation loss and a smooth $\ell_{1}$ distance as flow loss. The overall loss function is given by:
\begin{equation}
	\mathcal{L} = \frac{1}{T_\text{out}} \left\{\sum_{t=0}^{T_\text{out}-1}\gamma^{t} \left(\lambda_1 \mathcal{L}_\text{ce}(\hat{y}_{t}^\text{seg}, y_{t}^\text{seg}) + \lambda_2 \mathcal{L}_{\ell_{1}}(\hat{y}_{t}^\text{flow}, y_{t}^\text{flow})\right)\right\},
\end{equation}
with a future discount parameter $\gamma=0.95$ and balance factors $\lambda_1$ and $\lambda_2$ that are dynamically updated using uncertainty weighting \cite{Kendall2017MultitaskLU}.

\subsection{Instance Association}
\label{sec:instance}
\begin{figure}[t]
	\centering
	\includegraphics[width=0.9\linewidth]{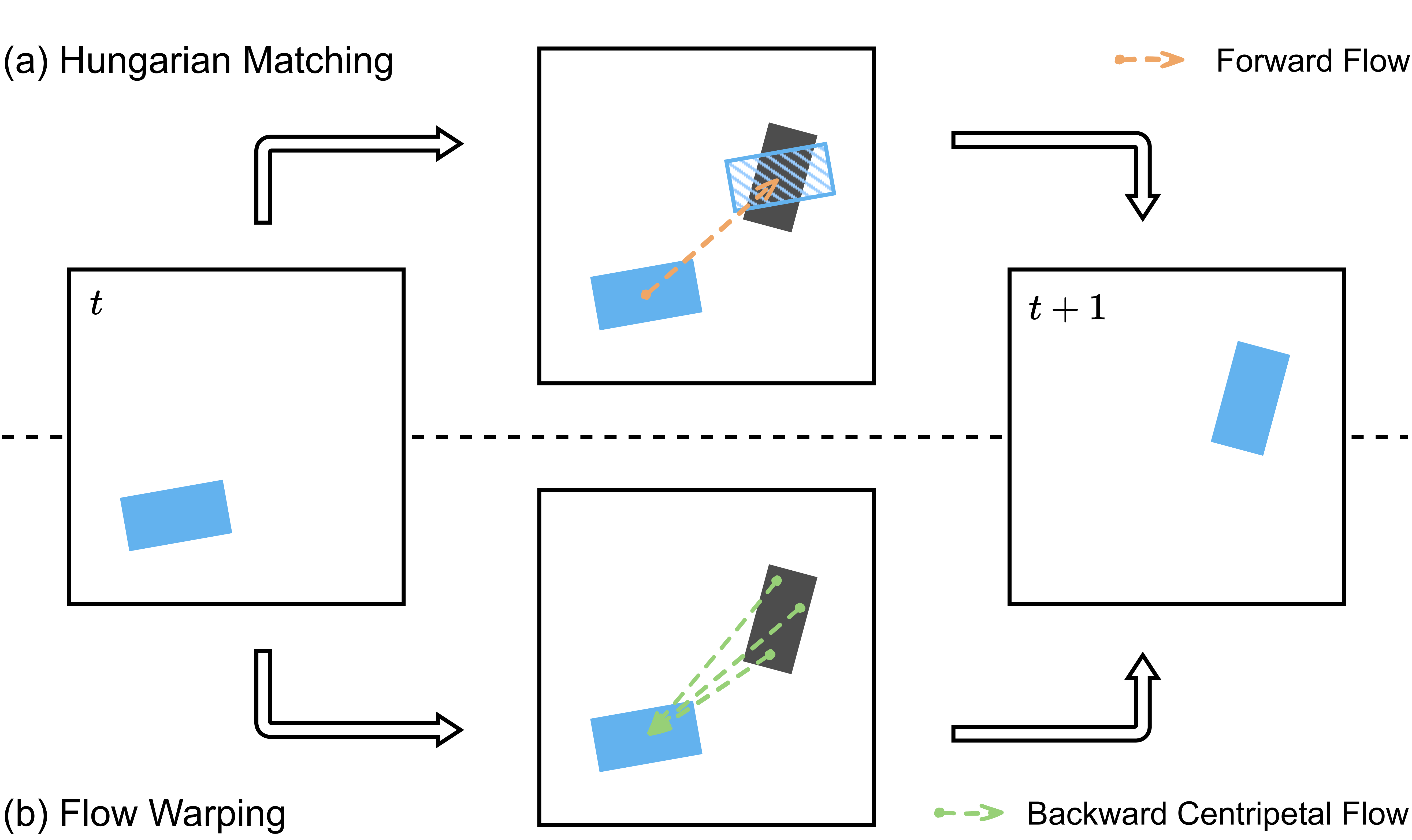}
	\caption{\textbf{Instance Matching Illustration:} The top branch is the Hungarian Matching algorithm (a) with forward flow as used in FIERY. The bottom branch is our backward flow warping operation (b) with centripetal backward flow.}
	\label{fig:postprocessing_illustrations}
	\vspace{-0.2cm}
\end{figure}

For the instance predictions, we need to associate the future instances $\left\{ \hat{y}_{t}^\text{inst} \right\}_{t=0}^{T_\text{out}-1}$ over time.
Existing methods project instance centers to the next frame using the forward flow, and then match the nearest agent centers using Hungarian Matching~\cite{Kuhn1955TheHM} as shown in~\figref{fig:postprocessing_illustrations}.a.
This method performs an \textit{instance-level association}, where an instance identity is represented by its center.
Therefore, only the flow vector located on the object center is used for motion prediction.
This has two disadvantages:
Firstly, object rotation is not considered and, secondly, a single displacement vector is more prone to errors than multiple displacement vectors covering the entire instance.
In practice, this can lead to overlapping projected instances, resulting in incorrect ID assignments.
This is particularly evident for close objects over a long prediction horizon.

Leveraging our backward centripetal flow, we further propose a \textit{warping-based pixel-level association} to tackle the above mentioned problems.
An illustration of our association method is shown in~\figref{fig:postprocessing_illustrations}.b.
For each foreground grid cell, this operation directly propagates the instance ID from the pixel at the flow vector destination in the previous frame to the current frame.
Using this method, the instance ID of each pixel is assigned seperately, yielding a pixel-level association.
Compared to the instance-level association, our method is tolerant to more severe flow prediction errors, because neighboring grid cells around the true center are inclined to share the same identity and errors tend to occur at individual peripheral pixels.
In addition, by using backward flow warping, multiple future positions can be associated with one pixel in the previous frame \cite{mahjourian2022occupancy}.
This is beneficial for the multi-modal future prediction.

As described, the backward association needs instance IDs at the previous frame.
A special case is the instance segmentation generation of the first frame ($t=0$), where no instance information at its previous frame ($t=-1$) is available.
Thus, only for the timestamp $t=0$, we assign instance IDs by grouping pixels to past instance centers.
This is similar to \cite{Cheng2019PanopticDeepLabAS} but without the usage of an additional centerness head since the centers are extracted from the semantic segmentation as discussed in~\secref{sec:multitask}.

%%%%%%%%%%%%%%%%%%%%%%%%%%%%%%%%%%%%%%%%%%%%%%%%%%%%%%%%%%%%%%%%%%%%%%%%%%%%%%%%
\section{Experimental Evaluation}
\label{sec:exp}

\subsection{Experimental Setup}
\label{sec:exp_setup}

\paragraph{Dataset}
We evaluate our method and compare it with state-of-the-art frameworks on the NuScenes Dataset \cite{Caesar2019nuScenesAM}, a widely used public dataset for perception and prediction in autonomous driving.
It contains 1\,000 driving scenes collected from Boston and Singapore, split into training, validation, and test sets with 750, 150, and 150 scenes, respectively.
Each scene consists of 20 seconds of traffic data and is labeled with semantic annotations at 2 Hz frequency.

\paragraph{Implementation Details}
We follow the setup from existing studies \cite{hu2021fiery,akan2022stretchbev,zhang2022beverse} that use the information of 3 frames corresponding to the past 1\,s (including the present frame) to predict the semantic segmentation, flow, and instance motion of 4 frames corresponding to the future 2\,s.
Input images are scaled and cropped to the size of $480 \times 224$, while the BEV map corresponds to a grid size of $200 \times 200$.
To evaluate the model performance at different perceptual scopes, two spatial resolutions are adopted: (1) $100$\,m $\times 100$\,m area with $0.5$\,m resolution (long) and (2) $30$\,m $\times 30$\,m area with $0.15$\,m resolution (short).
Using Adam optimizer with a learning rate of $3 \times 10^{-4}$, the end-to-end framework is trained for 20 epochs on four Tesla V100 GPUs with 16 GB memory and with a batch size of 8.
Our implementation is based on the code of FIERY \cite{hu2021fiery}.

\paragraph{Metrics}
We follow the evaluation procedure of FIERY \cite{hu2021fiery}.
To evaluate the segmentation accuracy, we use the Intersection-over-Union (IoU) as evaluation metric for the segmentation quality, i.e.:
\begin{equation}
	\operatorname{IoU}(\hat{y}_{t}^\text{seg}, {y}_{t}^\text{seg})=\frac{1}{T_\text{out}} \sum_{t=0}^{T_\text{out}-1} \frac{{\textstyle \sum_{h,w}} \hat{y}_{t}^\text{seg} \cdot {y}_{t}^\text{seg}}{{\textstyle \sum_{h,w} \hat{y}_{t}^\text{seg} + {y}_{t}^\text{seg} - \hat{y}_{t}^\text{seg} \cdot {y}_{t}^\text{seg}}},
\end{equation}
where $\hat{y}_{t}^\text{seg}$ and ${y}_{t}^\text{seg}$ are the predicted and ground truth semantic segmentations at timestamp $t$, respectively.
We use video panoptic quality (VPQ) as a more adequate metric for the instance prediction task. It consists of two parts: (1) recognition quality (RQ), i.e., the ID consistency of detected instances across the entire time horizon; (2) segmentation quality (SQ), i.e., the accuracy of the instance segmentation itself.
VPQ is calculated as
\begin{equation}
	\operatorname{VPQ}(\hat{y}_{t}^\text{inst}, {y}_{t}^\text{inst})=\sum_{t=0}^{T_\text{out}-1} \frac{ {\textstyle \sum_{(p_{t}, q_{t})\in TP_{t}} IoU(p_{t}, q_{t})}}{\left | TP_{t} \right | + \frac{1}{2} \left | FP_{t} \right | + \frac{1}{2} \left | FN_{t} \right |},
\end{equation}
where $TP_{t}$, $FP_{t}$ and $FN_{t}$ correspond to true positives, false positives and false negatives at time point $t$, respectively.

\paragraph{Baseline Methods}
We compare \name{} with the three state-of-the-art methods FIERY \cite{hu2021fiery}, StretchBEV \cite{akan2022stretchbev}, and BEVerse \cite{zhang2022beverse}.
FIERY and StretchBEV have the same experimental setup as our work, except for a larger batch size of 12 on four Tesla V100 GPUs with 32GB memory each.
BEVerse upgrades the backbone to the more advanced SwinTransformer \cite{liu2021swin}, significantly increasing the image input size to $704 \times 256$ and the batch size to 32 using 32 NVIDIA GeForceRTX 3090 GPUs to train the end-to-end model.
To demonstrate the effectiveness of our framework, we intentionally do not use large models or large image size like BEVerse, but limit ourselves to the FIERY setting in terms of FLOPs and GPU memory usage for a fair comparison.

\subsection{Label Generation Optimization}
In preliminary experiments, we found that the data pre-processing and label generation of the baseline method FIERY introduces systematic errors.
The original data pre-processing consists of three steps:
(1) Surrounding vehicle positions are transformed from the global coordinate system (GCS) to the ego coordinate system (ECS) at the corresponding timestamp based on the ego-motion.
At this timestamp, the instance parameters are then rendered into BEV maps to generate segmentation, instance, centerness and offset ground truth.
(2) The generated instance map is warped by ego-motion to obtain the warped instance map of the next frame.
The flow map is calculated as the geometric center displacement of the future warped instance from the current instance where all pixels of this instance share the same flow.
(3) During pre-processing, all BEV maps are transformed back into GCS.
As all these steps are performed in discrete BEV space, the two inverse coordinate transformations in (1) and (3) introduce considerable numerical errors into the ground truth maps.
Additionally, these errors further spread to the flow generation:
Even for a stationary vehicle, its BEV segmentation will slightly jitter over time and may also be assigned to non-zero flow values.
Overall, these errors in the ground truth generation significantly affect the prediction performance.

To solve the above problems, we modify the process in both training and validation as follows:
We propose to determine the perceptual area based on ego-vehicle position and render BEV maps directly in GCS to avoid errors caused by these two inverse transformations.
In addition, the flow ground truth is calculated using the instance maps of two adjacent frames without warping.
We also filter the flow ground truth by zeroing all values below a threshold to eliminate noise and artifacts of stationary vehicles.

%%%%%%%%%%%%%%%%%%%%%%%%
\subsection{Comparison to Baselines}

\begin{table}[t]
	\centering
	\small
	\setlength{\tabcolsep}{9pt}
	\begin{tabular}{lcccc}
		\toprule
		\multirow{2}{*}{Method} & \multicolumn{2}{c}{IoU} & \multicolumn{2}{c}{VPQ}                                 \\
		                        & Short                   & Long                    & Short         & Long          \\
		\midrule
		FIERY                   & 59.4                    & 36.7                    & 50.2          & 29.9          \\
		FIERY$\ddagger$ (repr.) & 58.3                    & 38.2                    & 48.2          & 30.9          \\
		StretchBEV              & 55.5                    & 37.1                    & 46.0          & 29.0          \\

		BEVerse$\dagger$        & 60.3                    & 38.7                    & 52.2          & 33.3          \\
		\midrule
		\name$\ddagger$         & \textbf{62.5}           & \textbf{39.3}           & \textbf{55.5} & \textbf{33.8} \\
		\bottomrule
	\end{tabular}
	\caption{Instance prediction benchmark results on NuScenes dataset. $\dagger$ uses a larger image size of $704 \times 256$, others use $480 \times 224$. Models with $\ddagger$ use our optimized label generation.}
	\label{tab:benchmarks}
\end{table}

\begin{table}[t]
	\centering
	\small
	\renewcommand\arraystretch{1}
	\setlength{\tabcolsep}{2pt}
	\begin{tabular}{c|cccc|cc|cc}
		\toprule
		\multirow{2}{*}{Model} & \multirow{2}{*}{Segmentation} & \multirow{2}{*}{Flow} & \multirow{2}{*}{Center} & \multirow{2}{*}{Offset} & \multicolumn{2}{c|}{IoU} & \multicolumn{2}{c}{VPQ}                                 \\
		                       &                               &                       &                         &                         & Short                    & Long                    & Short         & Long          \\
		\midrule
		$[\text{A}]$           & \checkmark                    & \checkmark            & \checkmark              & \checkmark              & 61.7                     & 39.2                    & 52.8          & 33.5          \\
		$[\text{B}]$           & \checkmark                    & \checkmark            & \checkmark              & -                       & 62.4                     & 39.2                    & 55.3          & 33.1          \\
		$[\text{C}]$           & \checkmark                    & \checkmark            & -                       & \checkmark              & \textbf{62.5}            & 39.2                    & 51.8          & 33.6          \\
		$[\text{D}]$           & \checkmark                    & \checkmark            & -                       & -                       & \textbf{62.5}            & \textbf{39.3}           & \textbf{55.5} & \textbf{33.8} \\
		\bottomrule
	\end{tabular}
	\caption{Comparison of different multi-task training setups with different heads used. Our method $[\text{D}]$ requires fewer heads than the typical baseline approach $[\text{A}]$ and yields best results.}
	\label{tab:ablation_multi-tasking_settings}
\end{table}\par

\begin{table}[t]
	\centering
	\small
	\setlength{\tabcolsep}{2.5pt}
	% \vspace{-0.1cm}
	\begin{tabular}{c|cc|cc|cc}
		\toprule
		\multirow{2}{*}{Model} & \multirow{2}{1.5cm}{\centering Post-processing} & \multirow{2}{1cm}{\centering Time Horizon} & \multicolumn{2}{c|}{IoU} & \multicolumn{2}{c}{VPQ}                                 \\
		                       &                                                 &                                            & Short                    & Long                    & Short         & Long          \\
		\midrule
		$[\text{E}]$           & Instance-level HM                               & 2\,s                                       & 60.0                     & 38.3                    & 46.9          & 31.9          \\
		$[\text{F}]$           & Pixel-level Warping                             & 2\,s                                       & \textbf{61.7}            & \textbf{39.2}           & \textbf{52.8} & \textbf{33.5} \\
		\midrule
		$[\text{G}]$           & Instance-level HM                               & 8\,s                                       & 48.8                     & 33.4                    & 39.6          & 27.2          \\
		$[\text{H}]$           & Pixel-level Warping                             & 8\,s                                       & \textbf{57.7}            & \textbf{34.8}           & \textbf{48.2} & \textbf{29.4} \\
		\bottomrule
	\end{tabular}
	\caption{Comparison of different post-processing methods.}
	% \vspace{-0.2cm}
	\label{tab:ablation_post-processing}
\end{table}

\begin{table}[t]
	\centering
	\small
	\setlength{\tabcolsep}{2pt}
	% \vspace{-0.1cm}
	\begin{tabular}{l|cc|cc}
		\toprule
		\multirow{2}{*}{Method}   & \multicolumn{2}{c|}{IoU} & \multicolumn{2}{c}{VPQ}                                 \\
		                          & Short                    & Long                    & Short         & Long          \\
		\midrule
		FIERY (reproduced)        & 58.3                     & 38.2                    & 48.2          & 30.9          \\
		FIERY + Our Contributions & \textbf{61.6}            & \textbf{38.9}           & \textbf{56.4} & \textbf{32.4} \\
		\bottomrule
	\end{tabular}
	\caption{Ablation study regarding generalization capabilities of our design. We applied our multi-task setup and post-processing to the FIERY RNN-based backbone, yielding significant improvements.}
	\label{tab:pipeline_comparison}
	% \vspace{-0.2cm}
\end{table}

We first compare the performance of our method to baseline frameworks in~\tabref{tab:benchmarks}.
We also reproduced FIERY \cite{hu2021fiery} with our proposed label generation method, c.f.~\secref{sec:exp_setup}, yielding improvements in the long-range domain, which is essential for the safety of autonomous vehicles.

In comparison with the baseline methods, our approach achieves significant improvements in terms of both evaluation metrics IoU and VPQ for both perceptual range settings.
For the long-range setting, \name{} outperforms the reproduced FIERY by 1.1\% IoU and 2.9\% VPQ.
Furthermore, \name{} performs better than BEVerse in all metrics, despite using a lower input image resolution and less parameters.
Compared to other baselines that introduce a stochastic process in their model, \name{} is a deterministic approach which is able to accomplish accurate forecasting.
This also shows the ability of the backward flow in capturing multi-modal futures \cite{mahjourian2022occupancy}.

\begin{figure*}[t]
	\centering
	\includegraphics[width=0.9\linewidth]{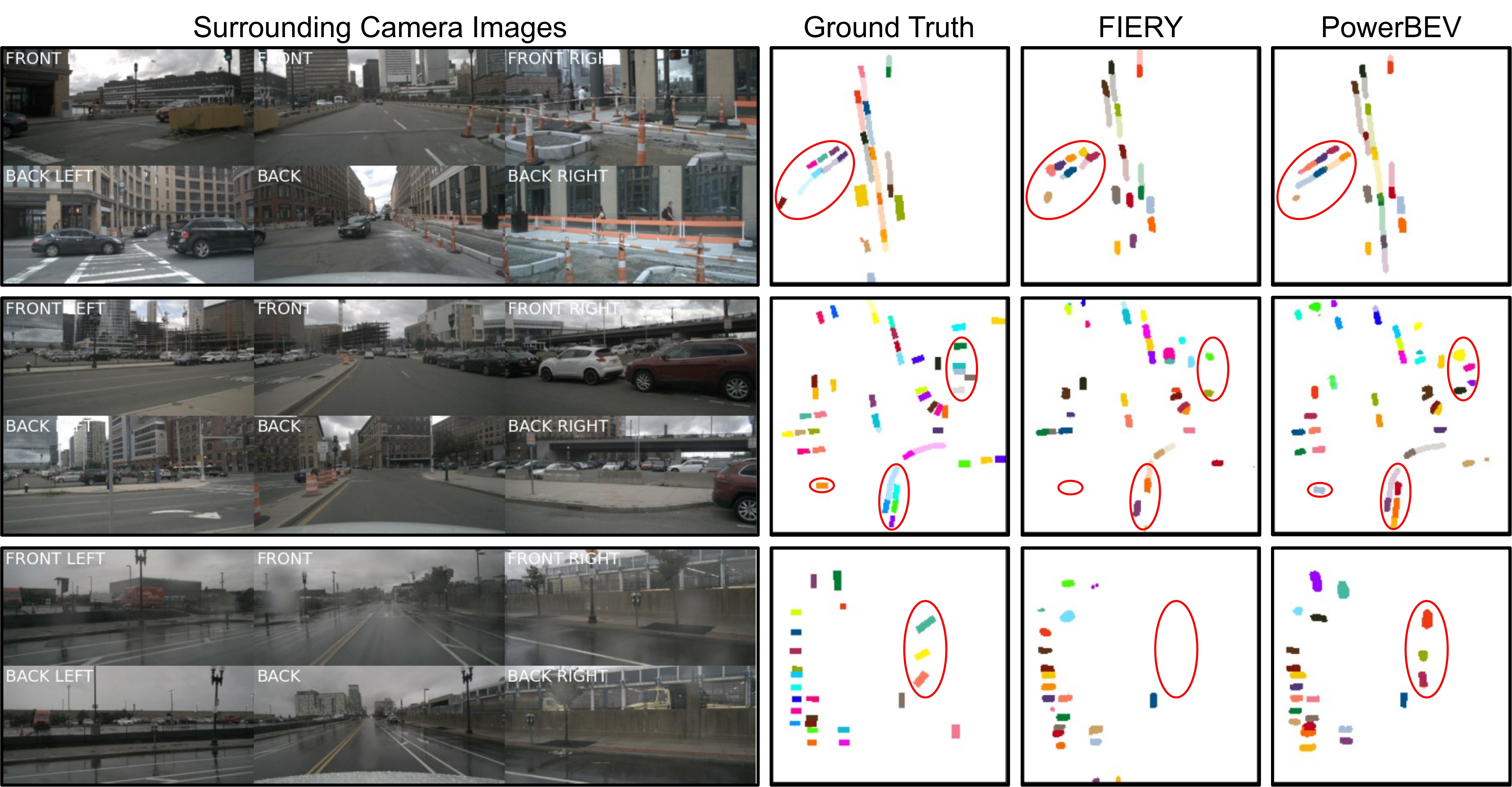}
	\caption{\textbf{Visualization of Instance Predictions:}
		We compare our approach with ground truth and the baseline FIERY.
		Each vehicle instance is assigned to a unique color and the predicted trajectory is represented by the same color with slight transparency.}
	\label{fig:visualization_comparison}
	% \vspace{-0.2cm}
\end{figure*}

\figref{fig:visualization_comparison} visualizes qualitative results of our method.
We show the comparison with FIERY in three typical driving scenarios: an urban scene with dense dynamic traffic, a parking lot with many static vehicles and a rainy scene.
Our approach provides more precise and reliable trajectory predictions for the most common dense traffic scenes, which becomes particularly evident in the first example with the vehicles turning into the side street on the left side of the ego-vehicle.
While FIERY only makes a few vague guesses about vehicle locations and has difficulties handling their dynamics, our approach, on the other hand, provides sharp object boundaries that match the real vehicle shapes better as well as their possible future trajectories.
Furthermore, as evident from the comparison in the second example, our framework can detect vehicles located at long distances, where FIERY fails.
In addition, our method detects the trucks occluded by walls in the rainy scene, which are difficult to spot even for human eyes.

%%%%%%%%%%%%%%%%%%%%%%%%
\subsection{Ablation Studies}
We conduct several ablation studies of \name{} to analyze the effectiveness of different components in our framework.

\subsubsection{Multi-task Learning}
We first investigate the effect of our reduced number of tasks and analyze whether typical problems in multi-task learning affect the performance, e.g. task balancing and training instability.
To this end, we keep the segmentation and backward centripetal flow heads as well as our post-processing unchanged, but attach additional heads to our prediction module (\secref{sec:msstc}). These heads are trained according to the objectives from the original FIERY baseline, i.e. centerness and centripetal offset.
The additional heads are trained jointly with segmentation and flow heads but not used in the post-processing.
All the experiments use uncertainty weighting \cite{Kendall2017MultitaskLU} to balance different tasks.

We vary the number and type of the additional training objectives, as shown in~\tabref{tab:ablation_multi-tasking_settings}.
Our approach with only two heads (Model $[\text{D}]$) performs better than all the other variants.
Adding the center (Model $[\text{B}]$) or offset (Model $[\text{C}]$) head negatively impacts various metrics.
For example, we observe a large decrease of VPQ for the short-range setting when training with the offset head.
The reason is that the backward centripetal flow is equal to the centripetal offset for static objects but different for moving objects, thus both training objectives lead to confusions during network training.
Compared to the Model $[\text{A}]$ that uses all the four heads as in existing works, Model $[\text{D}]$ achieves improvements of 2.7\% for the short and 0.3\% for the long-range setting in VPQ.
Another observation is that different loss terms converge at different speeds, making them difficult to support each other during training.
These findings support that eliminating redundant tasks is one of the sources for the performance improvements of our approach.
Although uncertainty weighting avoids tuning of loss weights, our approach directly reduces the amount of hyperparameters, which simplifies the balancing of different training objectives.

\subsubsection{Post-processing}
We further show the effectiveness of our proposed warping-based association for post-processing. To this end,
we want to compare our work with the traditionally used Hungarian Matching (HM), c.f. \figref{fig:motivation}.a.
As our network does not offer all required outputs for such a setup, we use the training setting of Model $[\text{A}]$ in~\tabref{tab:ablation_multi-tasking_settings} as a baseline.
As this model offers all four heads, we can directly compare both post-processing methods on the same network.
We also evaluate the prediction results over a longer prediction horizon (8s) to show the ability in maintaining ID consistency of different methods.

\begin{figure}[t]
	\centering
	\includegraphics[width=1.0\linewidth]{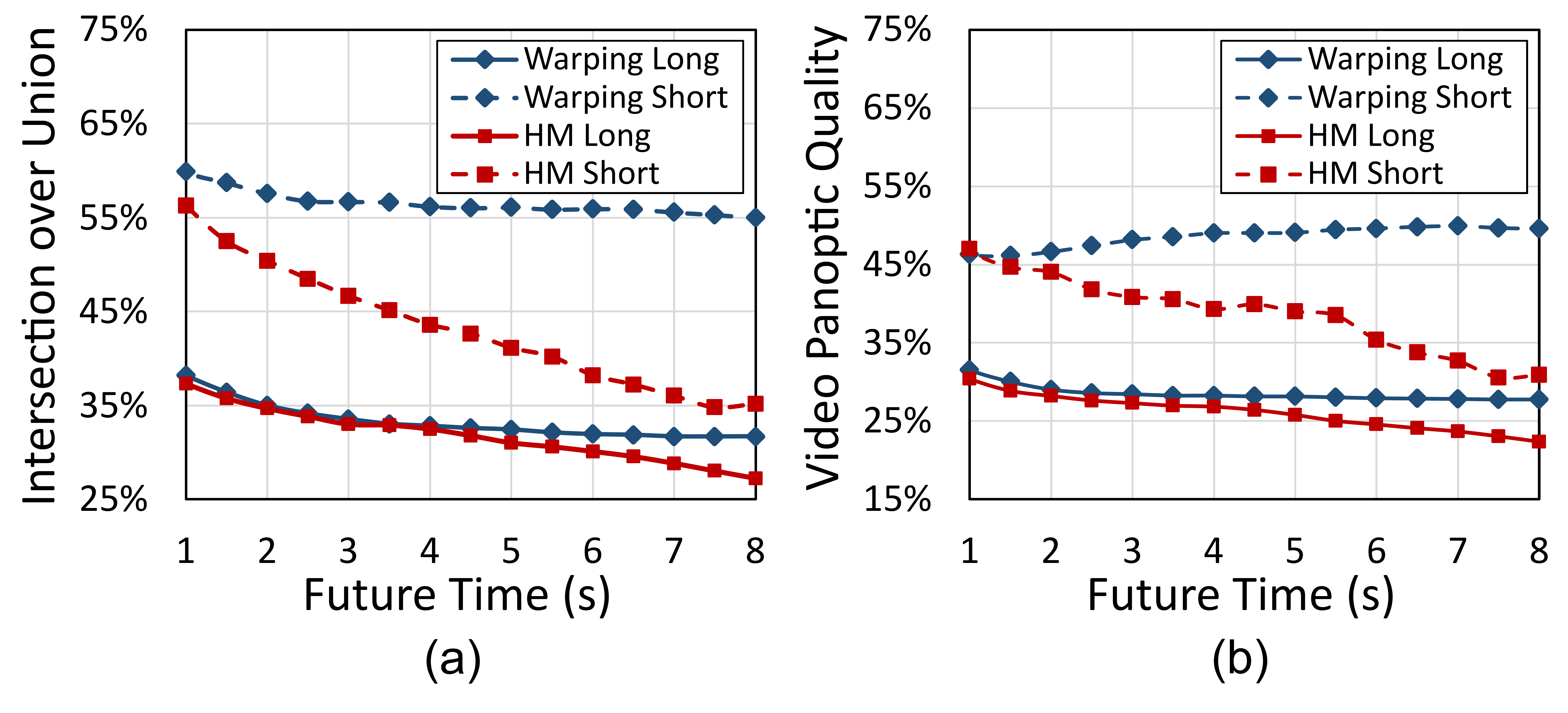}
	\vspace{-6mm}
	\caption{Comparison of different post-processing methods for the 8s time horizon: (a) IoU and (b) VPQ.}
	\label{fig:IoU_VPQ_comparison_8s_time_horizon}
\end{figure}

As evident from the the upper part of~\tabref{tab:ablation_post-processing}, our method (Model $[\text{F}]$) outperforms the HM-based instance-level association (Model $[\text{E}]$) in both IoU and VPQ.
For the longer time horizon, we observe an even more significant improvement of our method, especially for the short-range setting.
\figref{fig:IoU_VPQ_comparison_8s_time_horizon} provides a more detailed illustration of the metrics at each frame.
It can be observed that our flow warping remains stable over the long time horizon while the HM-based approach shows a clear performance decrease over time.
We attribute this to the fact that pixel-level association fully utilizes the network predictions of all pixels inside the instance boundary to boost the robustness.
In addition, our warping-based association is a many-to-one matching from the present to the past, which further increases temporal stability.

%%%%%%%%%%%%%%%%%%%%%%%%
\subsection{Generalization}

Next, we show that our design is not limited to CNN-based prediction models.
Our multi-task setting and flow warping operation can be ported to other model structures.
To verify this, we applied both to FIERY, keeping the remaining model structure and parameters unchanged.
The results in~\tabref{tab:pipeline_comparison} confirm that our approach also generalizes well to RNN-based models.
Hence, we believe that \name{} is a promising new paradigm for instance prediction and can serve as basis for future work.

%%%%%%%%%%%%%%%%%%%%%%%%%%%%%%%%%%%%%%%%%%%%%%%%%%%%%%%%%%%%%%%%%%%%%%%%%%%%%%%%
\section{Conclusion}
\label{sec:conclusion}

In this work, we presented a novel framework \bigname{} for future instance prediction in BEV.
Our approach only forecasts semantic segmentation and centripetal backward flow using 2D-CNNs in a parallel scheme.
It furthermore adopts a novel post-processing, which better handles multi-modal future motions, achieving state-of-the-art instance prediction performance in the NuScenes benchmark.
We provided thorough ablation studies that analyze our method and show its effectiveness. The experiments confirm that \bigname{} is more lightweight than previous approaches albeit yielding an improved performance. Hence, we believe that this method could become a new design paradigm for instance prediction in BEV.

%%%%%%%%%%%%%%%%%%%%%%%%%%%%%%%%%%%%%%%%%%%%%%%%%%%%%%%%%%%%%%%%%%%%%%%%%%%%%%%%
%% Future work: Use only if applicable -- but if so, use the following
%% sentence to start:
% Despite these encouraging results, there is further space for improvements.

%%%%%%%%%%%%%%%%%%%%%%%%%%%%%%%%%%%%%%%%%%%%%%%%%%%%%%%%%%%%%%%%%%%%%%%%%%%%%%%%
% Only if applicable
%\section*{Acknowledgments}
%We thank XXX for fruitful discussions and for \dots

\appendix
\section{Appendix}

\subsection{Framework Detail}

\paragraph{Pre-processing Stage and Perception Module:}
For the input image preprocessing and perception module structure, we follow the same setup as FIERY~\cite{hu2021fiery}.

\paragraph{Prediction Modules:}
We use two multiscale U-Net-like CNNs with identical structures to predict semantic segmentation and backward centripetal flow, respectively.
Each of these two branches has five scales from $200 \times 200$ to $7 \times 7$.

In the encoder, each scale has $N_{e}$-stacked encoder blocks ($N_{e}=3$) for spatio-temporal fusion of BEV features, followed by a downsampling layer with the stride of 2.
Each encoder block contains sequentially a 2D convolutional layer, a BatchNorm layer, and a LeakyReLU layer, together with an identity mapping.

Our predictor consists of five $N_{p}$-stacked predictor blocks ($N_{p}=5$) at five different scales.
The specific structure of the predictor block is the same as the encoder block and there is no feature communication between different scales.

The decoder has a mirror structure to the encoder, applying 2D transposed convolutional layers for upsampling.
The number of stacked decoder blocks is $N_{d}=3$.
In each step, the BEV features between different scales are concatenated and then decoded to a higher resolution.

At the end of each branch is a corresponding task head consisting of four Conv2D-BatchNorm-LeakyReLU blocks.
These convolution operations are only in the spatial dimension and irrelevant to the time.

\paragraph{Post-processing Stage:}
\name{} predicts the global vehicle motion of the future $T_\text{out}$ frames from the surround camera images by observing the past $T_\text{in}$ frames.
To address the special case of ID assignment in the first frame ($t=0$), the framework actually outputs an additional semantic segmentation for timestamp $t=-1$.
Following \cite{Cheng2019PanopticDeepLabAS}, we perform a max-pooling on the semantic segmentation at $t=-1$ to extract the local maxima as the centers of the instances.
The kernel size $k$ of max-pooling is flexibly adapted to the perceptual scope and spatial resolution: under short scope $k=7$; under long scope $k=23$ (i.e. $k$ is approximately equal to the size of a vehicle instance).
In addition, to prevent false positives, we apply a hard threshold of 0.1 to filter out low confidence center locations.

For future timestamps ($t>0$), \name{} uses \texttt{torch.grid\underline{ }sample} in combination with the backward centripetal flow for the occupied pixels in the semantic segmentation, obtaining the IDs of the corresponding previous positions.
Thus, future instance predictions are generated frame by frame in this warping fashion.

\begin{table*}[t!]
	\centering
	\small
	\caption{Runtime analysis of different stages in FIERY and \name{}.}
	\vspace{-0.1cm}
	\label{tab:runtime}
	\begin{tabular}{c|c|cc|C{1.6cm}C{1.6cm}C{2.2cm}C{1.2cm}}
		\toprule
		\multirow{2}{2cm}{\centering Method} & \multirow{2}{2.2cm}{\centering Time Horizon} & \multirow{2}{1.5cm}{\centering \#Param.} & \multirow{2}{1.5cm}{\centering  FLOPs} & \multicolumn{4}{c}{Runtime (ms)}                                        \\
		                                     &                                              &                                          &                                        & Perception                       & Prediction & Post-processing & Total \\
		\midrule
		FIERY                                & 2s                                           & 8.4M                                     & 206.8G                                 & 504                              & 36         & 82              & 622   \\
		\name{}                              & 2s                                           & 39.3M                                    & 92.6G                                  & 506                              & 46         & 14              & 566   \\
		\midrule
		FIERY                                & 8s                                           & 8.5M                                     & 709.5G                                 & 503                              & 62         & 396             & 961   \\
		\name{}                              & 8s                                           & 39.6M                                    & 108.4G                                 & 504                              & 52         & 59              & 615   \\
		\bottomrule
	\end{tabular}
	\vspace{-0.2cm}
\end{table*}

\subsection{Runtime Analysis}

\tabref{tab:runtime} compares the network parameters, FLOPs and average inference time of each stage between FIERY \cite{hu2021fiery} and our \name{} for two different prediction horizons: 2\,s and 8\,s.
We report the inference time on a compute node with an NVIDIA Tesla V100 GPU and 6 cores of an Intel Xeon E5-2690 v4 2.60GHz CPU.

Both FIERY and \name{} use the same perception module: a LSS-based \cite{philion2020lift} BEV feature extractor with spatial transformations based on ego-motion of different frames.
Thus, we observe no significant difference between each other in perception time.
Compared to other modules, the perception module significantly increases the latency.

In the 2s time horizon, our prediction module does not show a significant advantage over FIERY in terms of inference speed (46\,ms of \name{} vs.\ 36\,ms of FIERY).
This is because our multi-scale CNN architecture contains more learnable parameters (39.3M of \name{} vs.\ 8.4M of FIERY).
However, the RNN-based prediction approach used by FIERY requires recursive inference at each waypoint, which requires more FLOPs than \name{}.
As the time horizon is extended, this difference in FLOPs (709.5G of FIERY vs.\ 108.4G of \name{}) remarkably increases the prediction time of FIERY, which is 10\,ms longer than our method (52\,ms of \name{} vs.\ 62\,ms of FIERY).
This result shows the runtime advantage of the parallel prediction paradigm for a longer horizon compared to recursive or auto-regressive paradigms.

For both prediction horizons, our proposed post-processing runs about $6 \times$ faster than FIERY due to its simplicity.
Compared to the post-processing based on Hungarian Matching that is done on the CPU, our warping-based post-processing can be better deployed on the GPU.
Thus, our post-processing method has a higher potential to achieve faster speeds after implementation optimization.
In addition, the runtime of Hungarian Matching varies greatly for different numbers of agents, whereas our method maintains a better scalability due to stable runtime.

\subsection{Additional Visualization}

\begin{figure*}[ht]
	\centering
	\begin{subfigure}[h]{1.0\textwidth}
		\centering
		\includegraphics[width=0.95\textwidth]{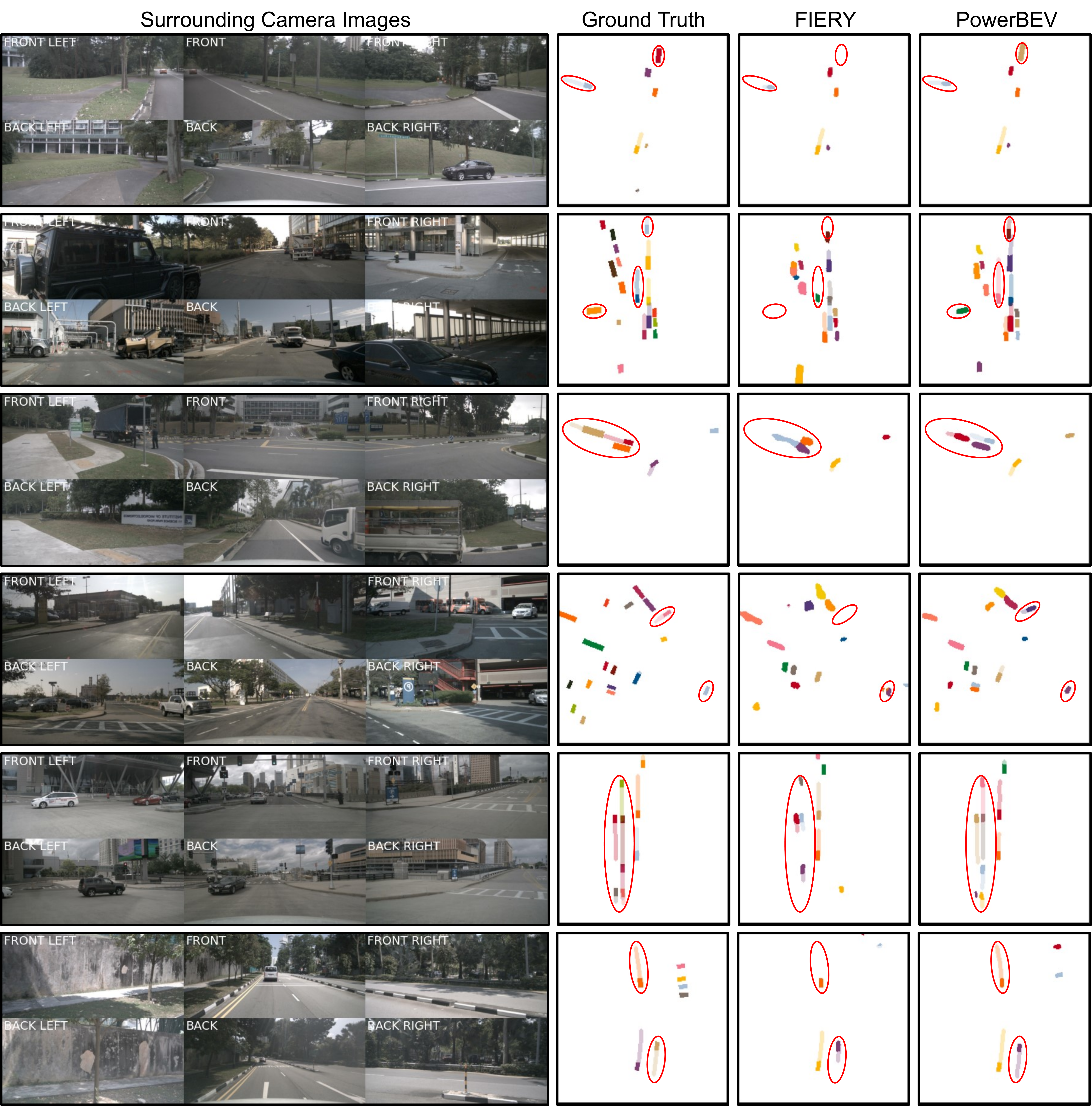}
		\caption{Daytime driving scenes.}
		\label{fig:a}
	\end{subfigure}
\end{figure*}
\begin{figure*}[t]
	\ContinuedFloat
	\centering
	\begin{subfigure}[h]{1.0\textwidth}
		\centering
		\includegraphics[width=0.95\textwidth]{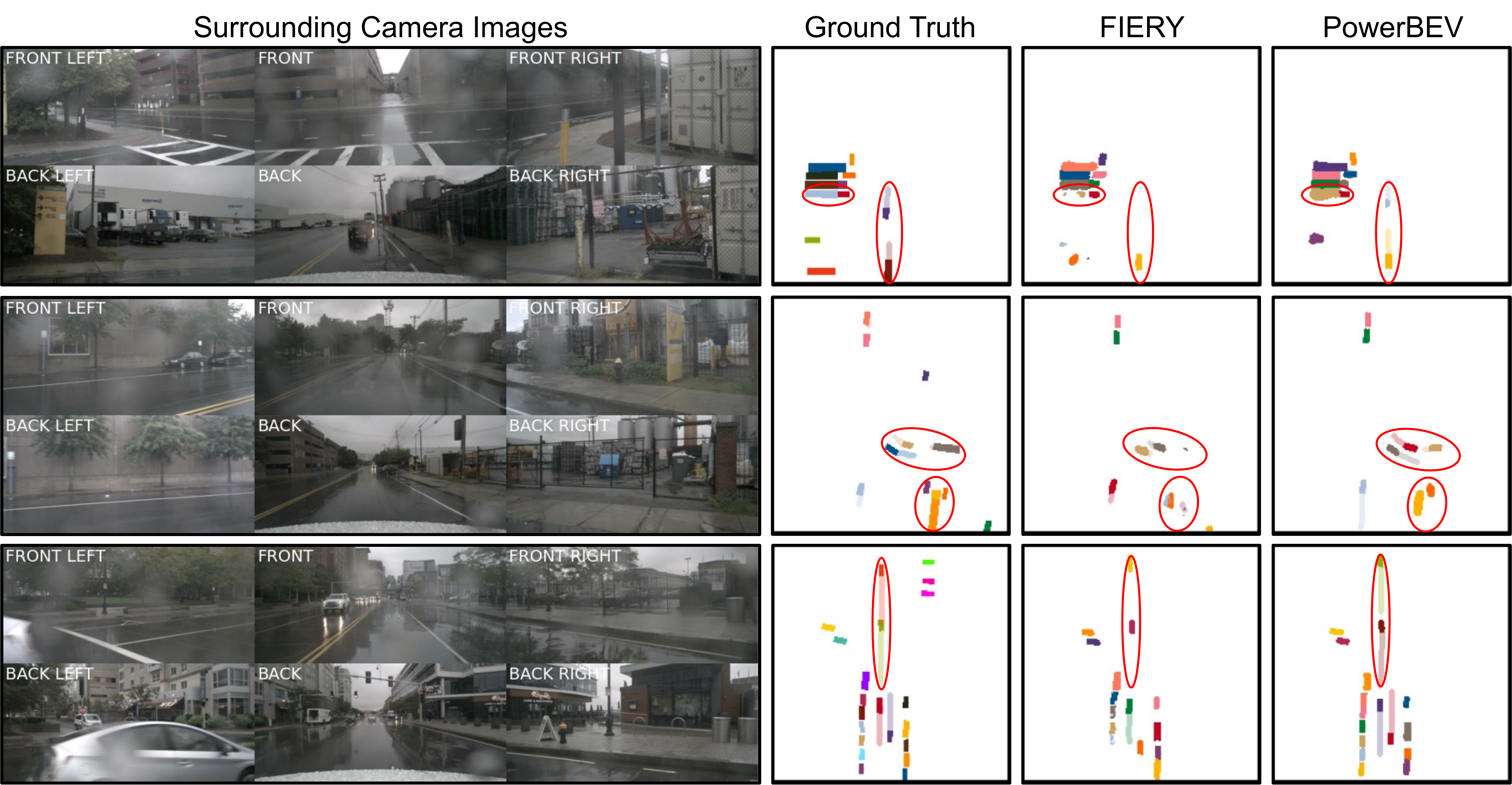}
		\caption{Rainy day scenes.}
		\label{fig:b}
	\end{subfigure}
\end{figure*}
\begin{figure*}[t]
	\ContinuedFloat
	\centering
	\begin{subfigure}[h]{1.0\textwidth}
		\centering
		\includegraphics[width=0.95\textwidth]{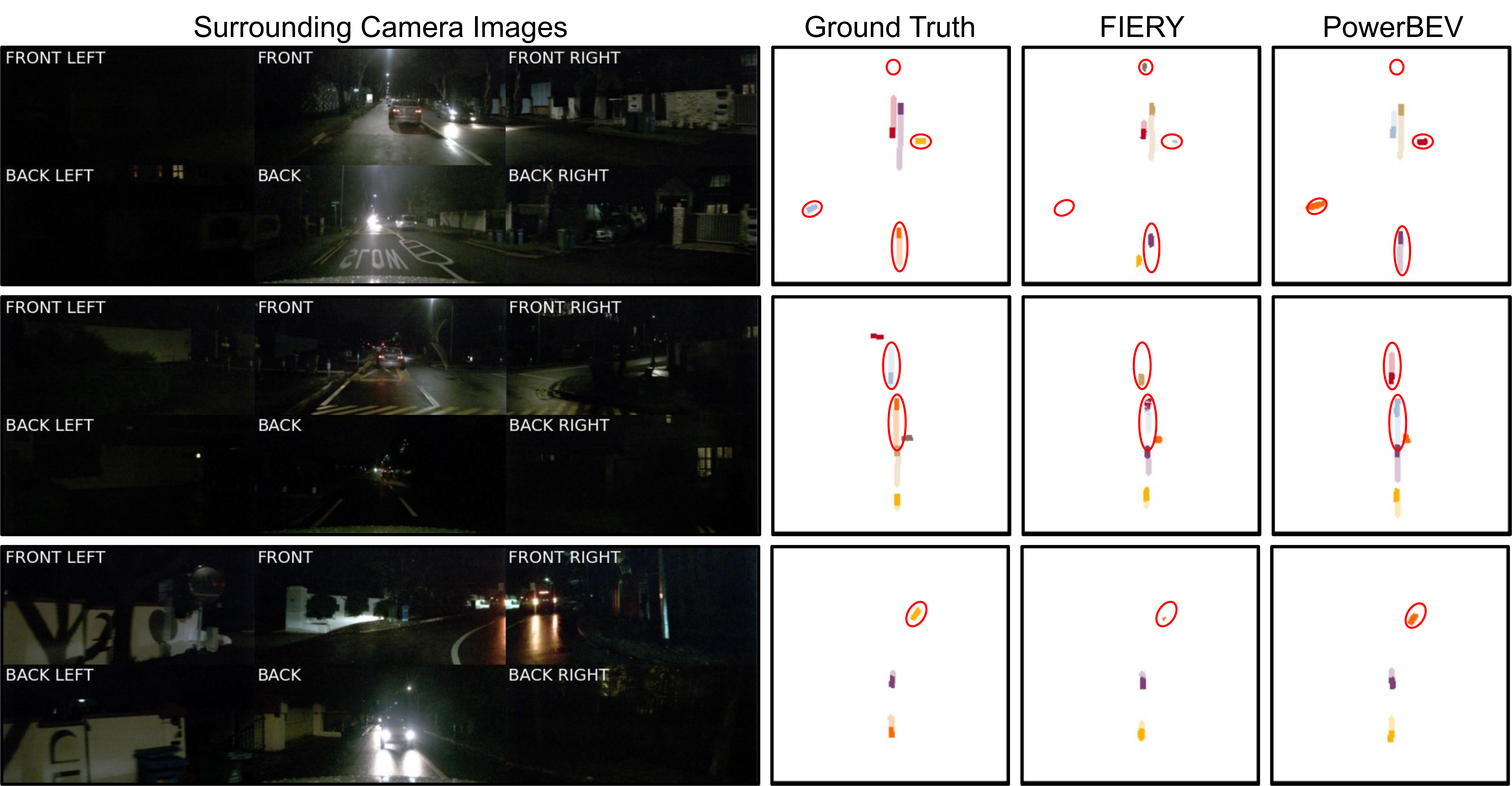}
		\caption{Night driving scenes.}
		\label{fig:c}
	\end{subfigure}
	\vspace{0.5cm}
	\caption{\textbf{Additional Visualization of Instance Predictions:}
		We provide more visualizations of our approach compared to ground truth and baseline FIERY.
		These include (a) daytime driving scenes, (b) rainy day scenes, and (c) night driving scenes.
		Each vehicle instance is assigned to a unique color and the predicted trajectory is represented by the same color with slight transparency.}
	\label{fig:app_vis}
	\vspace{-0.2cm}
\end{figure*}

To fully demonstrate the effectiveness of our framework, we further show additional qualitative comparisons in~\figref{fig:app_vis}.

As shown in~\figref{fig:app_vis}.a, our \name{} generates more accurate vehicle boundaries and better future motion trajectories on the most common daytime urban roads.
Compared to FIERY, our framework can still make reasonable predictions even when the vehicle is located at a long distance or partially occluded.

\figref{fig:app_vis}.b corresponds to the rainy day scenario.
Since some sight lines are blocked by raindrops, broken instances often appear in the prediction results of FIERY.
In contrast, the vehicles and trajectories predicted by our method show less artifacts.

\figref{fig:app_vis}.c shows the visualization comparison under the poor lighting condition at night.
This indicates that our framework can better avoid location estimation errors under bright light and vehicle prediction omissions in shadows.

\clearpage
\clearpage

\section*{Acknowledgments}

The research leading to these results is funded by the German Federal Ministry for Economic Affairs and Climate Action within the project “KI Delta Learning“ (Förderkennzeichen 19A19013A). The authors would like to thank the consortium for the successful cooperation.

Juergen Gall has been supported by the Deutsche Forschungsgemeinschaft (DFG, German Research Foundation) GA 1927/5-2 (FOR 2535 Anticipating Human Behavior) and the ERC Consolidator Grant FORHUE (101044724).

%% The file named.bst is a bibliography style file for BibTeX 0.99c
\bibliographystyle{named}
\bibliography{ijcai23}

\end{document}